\documentclass[sigconf,nonacm]{acmart}
\AtBeginDocument{%
  }

\usepackage{enumitem}
\usepackage{graphicx}
\usepackage{amsmath}
\usepackage{pdfpages}
\usepackage{makecell}
\usepackage{array}
\usepackage{tabularx}
\usepackage{stfloats} 
\usepackage{setspace}
\usepackage{comment}
\usepackage{colortbl}
\usepackage{booktabs}
\usepackage{appendix}
\usepackage{listings}
\usepackage{courier}
\usepackage{caption}
\usepackage{pgfplots,subcaption}
\usepgfplotslibrary{groupplots}
\usepackage{courier}
\usepackage{fix-cm}
\usepackage{siunitx}
\usepackage[most]{tcolorbox}
\usepackage{tcolorbox}
\usetikzlibrary{shadows}
\definecolor{HeaderGray}{gray}{0.85}
\definecolor{RowGray}{gray}{0.95}
\definecolor{HeaderPurple}{HTML}{D8CFF0}
\definecolor{RowLightPurple}{HTML}{F4F0FB}
\definecolor{BoxGray}{gray}{0.95}
\definecolor{LightOrange}{HTML}{FFD9B3}   
\definecolor{VeryLightOrange}{HTML}{FFF3E0}
\definecolor{finGray}{gray}{0.85}
\colorlet{finEv}{teal!45}
\colorlet{finCr}{orange!55!white}
\colorlet{finFull}{cyan!45!blue!35}
\definecolor{LimeGreen}{rgb}{0.196, 0.804, 0.196}

\newcolumntype{L}{>{\raggedright\arraybackslash}X}

\definecolor{lightpurple}{RGB}{230,230,250} 

\pgfplotsset{compat=1.18}

\lstdefinestyle{jsonstyle}{
    language=,
    basicstyle=\footnotesize\ttfamily,
    breaklines=true,
    showstringspaces=false,
    frame=single,
    backgroundcolor=\color[gray]{0.95}
}

\usepackage{hyperref}
\hypersetup{
  colorlinks   = true,
  citecolor    = Purple, 
  linkcolor    = MidnightBlue,
  urlcolor     = blue,
}
\usepackage[strings]{underscore}

\begin{document}

\title{A Role-Aware Multi-Agent Framework for Financial Education Question Answering with LLMs}


\author{Andy Zhu}
\affiliation{%
  \institution{Rensselaer Polytechnic Institute}
  \city{Troy, New York}
  \country{USA}
  }
\email{zhua6@rpi.edu}

\author{Yingjun Du}
\affiliation{%
  \institution{University of Amsterdam}
  \city{Amsterdam}
  \country{the Netherlands}
  }
\email{y.du@uva.nl}


\begin{abstract} 
Question answering (QA) plays a central role in financial education, yet existing large language model (LLM) approaches often fail to capture the nuanced and specialized reasoning required for financial problem-solving. The financial domain demands multi-step quantitative reasoning, familiarity with domain-specific terminology, and comprehension of real-world scenarios. We present a multi-agent framework that leverages role-based prompting to enhance performance on domain-specific QA. Our framework comprises a Base Generator, an Evidence Retriever, and an Expert Reviewer agent that work in a single-pass iteration to produce a refined answer. We evaluated our framework on a set of 3,532 expert-designed finance education questions from Study.com, an online learning platform. We leverage retrieval-augmented generation (RAG) for contextual evidence from 6 finance textbooks and prompting strategies for a domain-expert reviewer. Our experiments indicate that critique-based refinement improves answer accuracy by 6.6--8.3\% over zero-shot Chain-of-Thought baselines, with the highest performance from {Gemini-2.0-Flash}. Furthermore, our method enables {GPT-4o-mini} to achieve performance comparable to the finance-tuned {FinGPT-mt\_Llama3-8B\_LoRA}. Our results show a cost-effective approach to enhancing financial QA and offer insights for further research in multi-agent financial LLM systems. 
 \end{abstract} 
 




\maketitle


\section{Introduction}
Finance education is a rapidly expanding field in which students are expected to bridge the gap between financial literacy and industry application. Mastery requires proficiency in handling the tabular, textual, and numerical forms of structured financial data. Yet, even business school graduates often struggle to apply their knowledge to solve real-world business challenges. Professional certification exams like the Chartered Financial Analyst (CFA), Certified Public Accountant (CPA), and Financial Risk Manager (FRM), have been historically difficult to pass with roughly 40-50\% success rates. These exams not only test for knowledge recall but also the ability to translate complex financial principles to actionable solutions. The complexity of financial education extends well beyond basic data comprehension and into highly-specialized skills such as equity valuation, understanding derivatives on financial statements, and recognizing regulatory requirements. Large language models have shown a promising opportunity to democratize financial education through adaptive instruction and personalized feedback. For them to be effective, LLMs must demonstrate deep conceptual mastery across all finance topics and must confidently apply content from course materials.

\begin{figure}[t]
\centering
\includegraphics[width=\columnwidth]{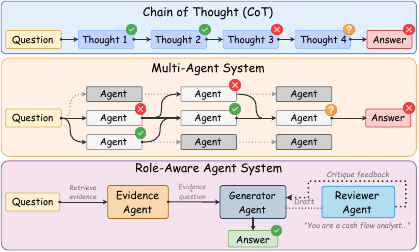} 
\captionsetup{skip=0.4pt}
\caption{Chain-of-Thought (CoT) uses forward reasoning with intermediate thoughts before arriving at an answer. Multi-Agent System uses multiple sequential agents but can propagate inconsistent reasoning. Our role-aware agent system for finance QA uses a dedicated Evidence agent for evidence, a Generator agent for answer drafting, and an Expert Reviewer agent for critique.}
\label{fig:intro_pipeline}
\Description{Chain-of-Thought (CoT) uses forward reasoning with intermediate thoughts before arriving at an answer. Multi-Agent System uses multiple sequential agents but can propagate inconsistent reasoning. Our role-aware agent system for finance QA uses a dedicated Evidence agent for evidence, a Generator agent for answer drafting, and an Expert Reviewer agent for critique.}
\end{figure}

Large language models (LLMs) have already demonstrated strong performance in general-domain reasoning tasks such as summarization, commonsense QA, and open-domain dialogue~\cite{brown2020language,openai2023gpt4}, where inputs are short and syntactically clean. Traditional (STEM) sciences benchmarks such as engineering and mathematics are typically narrow-scoped and have well-defined formulas that produce deterministic answers. Finance questions are particularly challenging because they are often embedded within longer contexts and unstructured inputs (e.g., tabular data, stock sheets, balance sheets). Correct answers depend on the skills of interpreting unstructured data, chaining multi-step numerical calculations, and adapting to evolving regulations and market dynamics. Even largely fine-tuned models trained on financial data often underperform on finance-QA tasks~\cite{wu2023bloomberggpt,huang2024open}. 

Recent advances in role prompting have shown that assigning expert personas to LLMs can steer their reasoning behaviors and improve alignment with domain-specific tasks~\cite{zheng2023helpful,kim2024persona}. In early studies, Chain-of-Thought prompting, as shown in Figure~\ref{fig:intro_pipeline}, uses the model's own internal representations to generate free-form thoughts. However, these thoughts can become ungrounded, limiting the model's ability to reason and iteratively update its knowledge. More recently, multi-agent systems have combined verbal reasoning with interactive decision making to create more autonomous systems~\cite{yao2023react}. Multi-agent systems in question answering were proposed to combine the roles of several specialized agents to vote and reach a better answer~\cite{kaesberg2025voting}. However, most frameworks use numerous refinement agents, which dilute context windows and raises latency and token costs. Passing between ``until satisfaction" loops exacerbates these issues and also makes it difficult to trace which step produced an error. 

Financial analysts, auditors, and researchers work in tandem to decompose financial tasks. Financial QA contains multi-faceted subjects where factual evidence and domain-specific expertise are crucial to produce the most-grounded results. In this work, we propose a role-aware multi-agent prompting framework tailored for financial question answering. The system comprises 3 agents: a Base Generator that performs step-by-step reasoning and answer generation, an Evidence Retriever that retrieves topic-relevant evidence, and an Expert Reviewer that audits previous answers and produces a thorough critique for refinement. These agents are guided by topic-specific role prompts (e.g., tax analyst, equity expert) to enable confident domain-aligned behavior. Figure~\ref{fig:intro_pipeline} summarizes the differences between conventional Chain-of-Thought reasoning, generic multi-agent pipelines, and our proposed role-aware agent system. 

We evaluate 4 configurations of our multi-agent framework by testing 4 out-of-the-box LLMs--GPT-4o-mini, Gemini-2.0-Flash, Llama-3.1-70B-Instruct, and a finance-tuned baseline FinGPT-mt\_Llama3-8B\_LoRA--across the 82 financial topic question sets from Study.com. We group the topics under 7 more generalized finance categories and analyze the performance enhancements. Results show that implementing feedback from the Expert Reviewer agent consistently improve both answer accuracy and explanation quality over standard zero-shot Chain-of-Thought prompting. Incorporating contextual evidence from the Evidence Retriever agent further improves performances by a smaller margin than single agent frameworks. Using carefully crafted role prompts and reference material, we demonstrate that even out-of-the-box LLMs can achieve performance comparable to that of a fine-tuned LLM. 

\section{RELATED WORK} 

\subsection{Advancements in Financial QA} 
Early financial-QA studies relied on few-shot prompting with Chain-of-Thought examples to enable step-wise reasoning from general-domain LLMs. Chen et al.~\cite{chen2022convfinqa} explored in-context CoT exemplars in financial QA and showed that even with carefully chosen prompts few-shot CoT only yields under 50\% execution `accuracy for {GPT-3}. Subsequent refinements such as \textit{self-consistency} and dynamic N-shot prompting which adapt the exemplar set to each question's form and arithmetic depth, improved performance by 4\%-7\%. Chen et al.~\cite{chen2022program} introduced ``Program-of-Thoughts (PoT)" which lets the LLM emit short Python snippets that an external executor runs, effectively separating symbolic reasoning from floating-point computation. PoT beats CoT by 12\% on financial QA benchmarks. Retrieval-augmented generation (RAG) aims to chunk and index lengthy SEC filings and analyst reports~\cite{lai2024sec}. FinSage pairs hierarchical chunking with a numerical-aware re-ranker and cuts hallucinations in SEC-QA by 35\%~\cite{wang2025finsage}. 

\subsection{Role-Guided Reasoning in LLMs}
Instructing Large Language Models (LLMs) to assume specific roles and personas has shown to be effective for reasoning alignment and task performance. Role-guided prompting can simulate domain-specific expertise, making models adopt structured thinking paths and inference styles~\cite{xu2023expertprompting}. Kong et al.~\cite{kong2023better} proposed a two-stage role-play prompting that improves zero-shot performance across 12 general reasoning benchmarks covering arithmetic, commonsense, and symbolic tasks. Their method increased {GPT-3.5-turbo's} accuracy on the AQuA dataset from 53.5\% to 63.8\% and outperformed standard zero-shot-CoT on 9 out of 12 datasets. Additionally, Zheng et al.~\cite{zheng2023helpful} found that minor changes in role prompts significantly influenced models' intermediate reasoning paths. Kang et al.~\cite{kang2025exploring} introduced a critical distinction between Professional-Based Personas (PBPs), such as “scientist” or “accountant”, which invokes domain expertise, and \textit{Occupational Personality-Based Personas} (OPBPs), such as “a scientific person”, which reflects cognitive traits rather than knowledge competence. These studies suggest that carefully crafted expert role prompts can act as implicit triggers for multi-step reasoning.


\begin{figure*}[t]
\centering
\includegraphics[width=1.0\textwidth]{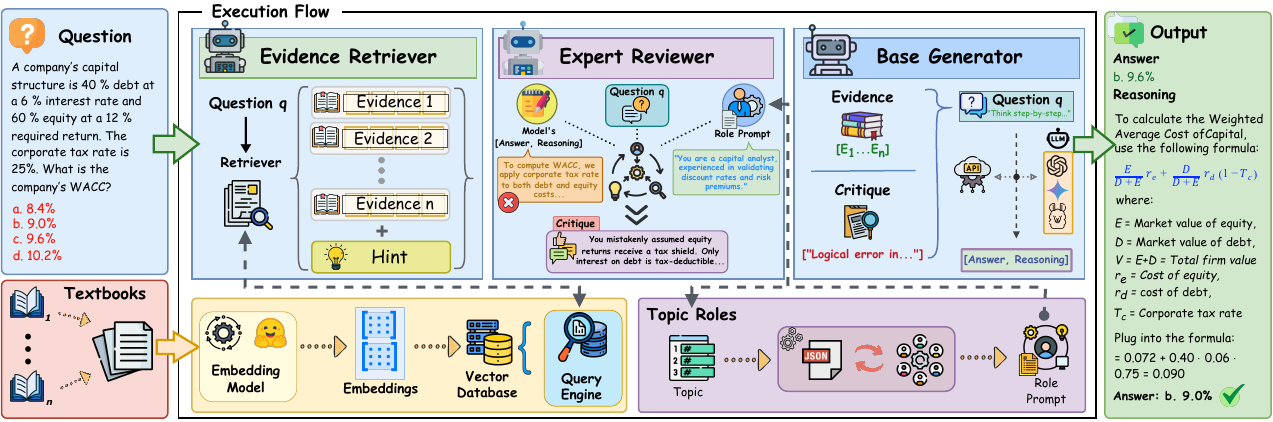} 
\captionsetup{skip=1pt}
\caption{Overall architecture of the role-aware financial QA pipeline. The framework consists of 3 core agents: the \textbf{Evidence Retriever} retrieves topic-relevant evidence from textbook-derived databases; the Expert Reviewer assesses the draft answer and injects expert feedback based on topic-specific role prompts; and the Base Generator composes the final answer by integrating retrieved evidence, critique, and the original query.}
\label{fig:pipeline_structure}
\Description{Overall architecture of the role-aware financial QA pipeline. The framework consists of 3 core agents: the \textbf{Evidence Retriever} retrieves topic-relevant evidence from textbook-derived databases; the Expert Reviewer assesses the draft answer and injects expert feedback based on topic-specific role prompts; and the Base Generator composes the final answer by integrating retrieved evidence, critique, and the original query.}
\end{figure*}

\begin{figure*}[t]
    \begin{tcolorbox}[
      colback=VeryLightOrange!100, 
      colframe=LightOrange!150, 
      arc=1.5pt, 
      outer arc=2pt,
      boxrule=0.5pt,
      width=\textwidth, 
      top=2pt,
      bottom=2pt,
      fontupper=\small,
      leftrule=0.5pt, rightrule=0.5pt, toprule=0.5pt, bottomrule=0.5pt, 
      title=\textbf{\hspace{0.15em} Topic \hspace{7.9em} Role Prompt},
      fonttitle=\large\bfseries,
      coltitle=black,
      boxsep=3.5pt,
      top=3pt,
      bottom=3pt,
      before skip=3pt,
      after skip=3pt,
    ]
    \small
      \setlist[description]{
        style=multiline,     
        leftmargin=4cm,      
        labelwidth=3.5cm,    
        labelsep=1em,        
        itemsep=2.2pt,
        align=left,          
        font=\normalfont,
        afterlabel=\mbox{}\\[2pt]
      }
      \begin{description}
        \item[\textbf{Dividend}]            You are a dividend‑policy expert, proficient in dividend strategies and payout analysis.
        \item[\textbf{Bonds}]               You are a bond‑market expert with deep knowledge of fixed‑income valuation.
        \item[\textbf{Budget}]              You are a budgeting specialist who designs and analyzes business budgets.
        \item[\textbf{Corporate Finance}]   You are a corporate‑finance strategist skilled in capital‑structure decisions.
        \item[\textbf{Credit}]              You are a credit analyst, experienced in evaluating credit risk and borrower quality.
        \item[\textbf{Derivatives}]         You are a derivatives specialist, versed in options, futures, and hedging techniques.
        \item[\textbf{Financial Risk}]      You are a financial‑risk manager who identifies and mitigates market and credit risk.
        \item[\textbf{Financial Statements}] You are a financial‑statement analyst adept at assessing firm performance.
        \item[\textbf{Investments}]         You are an investment analyst skilled in evaluating diversified investment opportunities.
        \item[\textbf{Portfolio Management}]You are a portfolio manager, expert in constructing and rebalancing portfolios.
        \item[\textbf{Stock Valuation}]     You are a stock‑valuation expert focused on intrinsic‑value estimation.
        \item[\textbf{.....}] .....
      \end{description}
    \end{tcolorbox}
    \caption{Sample role prompts used by the Expert Reviewer Agent.}
    \label{tab:role_prompts}
    \Description{Sample role prompts used by the Expert Reviewer Agent.}
\end{figure*}

\subsection{Finance-specific LLMs and Benchmarks}
Bloomberg introduced BloombergGPT~\cite{wu2023bloomberggpt}, a 50-billion parameter model trained on financial data that achieves strong results on sentiment analysis and named entity recognition. Similarly, OpenFinLLMs~\cite{huang2024open} released a suite of open-source multimodal LLMs (e.g., FinLLaMA) optimized for financial document understanding. FinGPT~\cite{yang2023fingpt} is an open-source family of {Llama-3-8B} fine-tuned on multi-task data from financial news and reports. FinGPT is designed for further fine-tuning with specific use cases like robo-advising and algorithmic trading. While these models demonstrate improved representation capacity, they remain resource-intensive and lack thorough QA evaluation. Benchmarks like FinQA~\cite{chen2021finqa} and FinBen~\cite{xie2024finben} focus on general applied reasoning skills useful for evaluating industry-capabilities but not always central to educational learning objectives. FinTextQA~\cite{chen2024fintextqa}, on the other hand, blends textbook questions on personal budgeting with government policies into one undifferentiated test set. 
Overall, many of these datasets target data extraction and multi-modal interpretation, overlooking educationally-grounded conceptual and arithmetic questions. These datasets generally fall short with limited topic diversity expected in finance education curriculum.


\begin{figure*}[t]
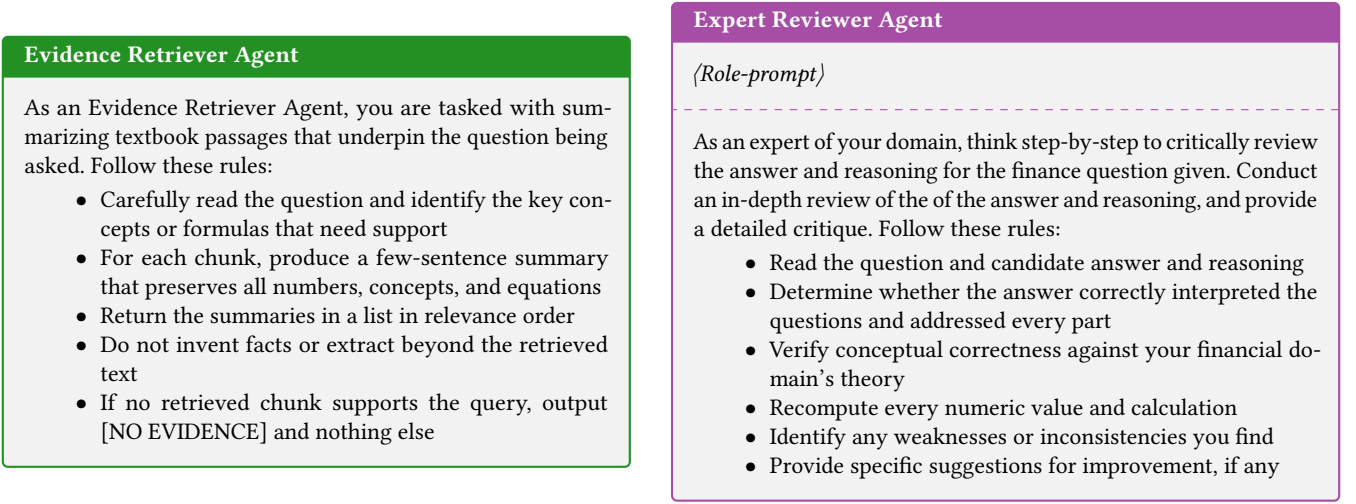

  \centering
  \begin{minipage}{0.47\textwidth}
    \begin{tcolorbox}[
      title=\textbf{Evidence Retriever Agent},
      colback=gray!10,
      coltext=black, 
      colframe=LimeGreen!70!black,
      fonttitle=\color{white},
      boxrule=0.8pt,
      arc=1.5pt,
      left=5pt,
      right=5pt,
      top=5pt,
      bottom=5pt
    ]
    
    As an Evidence Retriever Agent, you are tasked with summarizing textbook passages that underpin the question being asked. Follow these rules: 
    
    \begin{itemize}
      \item Carefully read the question and identify the key concepts or formulas that need support
      \item For each chunk, produce a few-sentence summary that preserves all numbers, concepts, and equations
      \item Return the summaries in a list in relevance order
      \item Do not invent facts or extract beyond the retrieved text
      \item If no retrieved chunk supports the query, output \hspace{1em} [NO EVIDENCE] and nothing else
    \end{itemize}
    \end{tcolorbox}
  \end{minipage}
  \hfill
  \begin{minipage}{0.50\textwidth}
    \begin{tcolorbox}[
      title=\textbf{Expert Reviewer Agent},
      colback=gray!10,           
      colframe=violet!70,
      fonttitle=\color{white},
      boxrule=0.8pt,
      arc=1.5pt,
      left=5pt,
      right=5pt,
      top=5pt,
      bottom=5pt
    ]
\textit{⟨Role-prompt⟩}
  \tcblower                                        
  As an expert of your domain, think step-by-step to critically review the answer and reasoning for the finance question given. Conduct an in-depth review of the of the answer and reasoning, and provide a detailed critique. Follow these rules: 
    \begin{itemize}
      \item Read the question and candidate answer and reasoning
      \item Determine whether the answer correctly interpreted the questions and addressed every part
      \item Verify conceptual correctness against your financial domain's theory 
      \item Recompute every numeric value and calculation
      \item Identify any weaknesses or inconsistencies you find 
      \item Provide specific suggestions for improvement, if any  
    \end{itemize}
    \end{tcolorbox}
  \end{minipage}
  \captionof{figure}{System prompts for the Base Generator (left) and Expert Reviewer agents (right).}
  \label{fig:agent_prompts}
  \Description{System prompts for the Base Generator (left) and Expert Reviewer agents (right).}
\end{figure*}

\section{METHODOLOGY}
We propose a role-aware multi-agent framework tailored for financial QA, designed to simulate expert reasoning through structured collaboration among agents, as shown in Figure~\ref{fig:pipeline_structure}.

\subsection{Role Prompting in Finance}
Finance tasks involve specialized knowledge and domain-specific terminology. Generic prompts often fail to activate domain-relevant reasoning strategies and mental models in LLMs. Prior work has shown that instructing models with role personas (e.g., “tax advisor” or “investment strategist”) can trigger more grounded, context-aware reasoning behaviors~\cite{zheng2024helpful,kong2023better}. For example, a model prompted with ``You are a cash-flow analyst, adept at interpreting liquidity statements and cash-flow mechanics” is likely to prioritize tasks such as evaluating operating versus finance cash flow. It may focus on detecting inconsistencies in cash reporting and trace differences in income and liquidity. A generic prompt may elicit answers that lack quantitative insights, especially within the niche domains like finance. 

While role prompting alone does not turn LLMs into financial experts, it can be a cost-effective method to steer reasoning in high-stakes, knowledge-intensive domains~\cite{wu2023bloomberggpt,huang2024open}. In our framework, each agent is assigned a domain-specific role in finance (Figure~\ref{tab:role_prompts}), enabling a model to prioritize financial concept reasoning over general reasoning. These benefits help the model simulate expert behavior and maintain coherence through its reasoning process, especially in complex financial contexts.

\subsection{Role-Aware Agent Framework}
The framework comprises 3 agents--Base Generator, Evidence Retriever, and Expert Reviewer--with each having a vital function in the question-answering process.

\begin{description}[leftmargin=0cm, labelsep=0.5cm]
\item[\textbf{Base Generator (BG)}]  
The Base Generator represents the default reasoning capability of an LLM. It queries an input question $q$ for an answer and reasoning through Chain-of-Thought reasoning. This underscores the model's out-of-the-box, zero-shot performance and can optionally incorporate external context $c$ from other agents: 

\[
(a^{(0)}, r^{(0)}) = \mathrm{BG}(q, c; \pi_{\text{cot}}), \quad c \in \{\varnothing, \mathcal{E}, \sigma\}
\]

The context $c$ may include evidence $\mathcal{E}$ or critique feedback $\sigma$ to be passed to the LLM along with the question. This gives additional context for the model to make enhanced answers. 

\item[\textbf{Evidence Retriever (ER)}]  
The Evidence Retriever simulates open-book evidence retrieval through retrieval-augmented generation (RAG). Given a query $q$, it finds the top-$k$ most semantically relevant chunks $[e_1, \dots, e_k]$ of text through a cosine similarity search. The agent then is then instructed (Figure~\ref{fig:agent_prompts}) to summarize the queried evidence into key facts: 

\[
\mathcal{E} = [e_1, \dots, e_k] = \text{ER}_k(q), \quad c = h \oplus \bigoplus_{i=1}^{k} e_i
\]

We also prepend a brief hint $h$ (e.g., “WACC is the weighted average cost of capital”) for each question for concept clarification. The retrieved evidence and hint are then concatenated into context block $c$.

\item[\textbf{Expert Reviewer (XR)}]  
The Expert Reviewer agent reviews the initial answer $(a^{(0)}, r^{(0)})$ produced by the Base Generator, identifying logical inconsistencies and conceptual misunderstandings.\footnote{Role prompting is applied only to the \emph{Expert Reviewer} to maintain a clean baseline where changes in accuracy can be attributed to evidence and/or critique rather than to an ``expert" prompting style. This also avoids anchoring bias~\cite{lou2024anchoring} and keeps inference costs low.} The agent is prompted (Figure~\ref{fig:agent_prompts}) as a domain-expert with a topic-specific role $t \in \mathcal{T}$:

\[
\sigma = \mathrm{XR}_t(q, a^{(0)}, r^{(0)}, c), \quad c \in \{\varnothing, \mathcal{E}\}
\]

Here, the Expert Reviewer checks the answer's strengths and weaknesses and produces a critique $\sigma$. The context $c$ may include the evidence from the Evidence Retriever to make a more informed evidence-backed critique.

\end{description}


\subsection{Testing Configurations}\label{sec:pipeline}
To examine the effects of retrieval-based evidence and expert critique in the system, we wire the 3 agents across 4 configurations, labeled \textit{M0} through \textit{M4}. Each mode incrementally enables the Evidence Retriever and Expert Reviewer. 

\paragraph{\textbf{M-0 (Closed-book)}}
\[
(q; \pi_{\text{cot}}) \xrightarrow{\text{BG}} (a, r)
\]
This mode evaluates the baseline reasoning capabilities of the model using the Base Generator without retrieved evidence or critique. It reflects the LLM’s zero-shot reasoning ability and produces an answer-reasoning pair. 

\paragraph{\textbf{M-1 (Evidence)}}
\[
q \xrightarrow{\text{ER}} \mathcal{E}, \quad (\mathcal{E}, q) \xrightarrow{\text{BG}} (a, r)
\]
This mode evaluates the effect of adding retrieved evidence. The Evidence Retriever provides evidence $\mathcal{E}$ from textbooks and the Base Generator consumes the evidence to produce an evidence-based answer-reasoning. 

\paragraph{\textbf{M-2 (Critique)}}
\begin{align*}
q &\xrightarrow{\text{BG}} (a^{(0)}, r^{(0)}) \tag*{(1)} \\
(q, a^{(0)}, r^{(0)}) &\xrightarrow{\text{XR}} \sigma,\quad (q, \sigma)\xrightarrow{\text{BG}} (a^{(1)}, r^{(1)}) \tag*{(2)}
\end{align*}

The Expert Reviewer audits the Base Generator's initial answer-reasoning and returns a critique $\sigma$. The Base Generator considers feedback $\sigma$ in a second inference pass, producing a revised answer-reasoning. 

\paragraph{\textbf{M-3 (Evidence + Critique)}}
\begin{align*}
q &\xrightarrow{\text{ER}} \mathcal{E} \xrightarrow{\text{BG}} (a^{(0)}, r^{(0)}) \tag{1}\\
(q, a^{(0)}, r^{(0)}, \mathcal{E}) &\xrightarrow{\text{XR}} \sigma \xrightarrow{\text{BG}} (a^{(1)}, r^{(1)}) \tag{2}
\end{align*}
The full role-aware agent pipeline combines retrieval-grounded reasoning with critique refinement. Evidence $\mathcal{E}$ from the Evidence Retriever is also passed to the Expert Reviewer along with the question. The LLM is invoked 4 times and outputs a final answer-reasoning. This represents our most robust setting and an example is illustrated in Figure~\ref{fig:pipeline_structure}. 


\begin{table}[t]
  \centering
  \scriptsize
  \setlength{\tabcolsep}{4pt}
  \begin{tabularx}{\linewidth}{@{}p{3cm} p{0.9cm} X@{}}
    \toprule
    \textbf{Category} & \textbf{Question Count} & \textbf{Topics}\\
    \midrule
    \textbf{1. Investments \& Valuation} \hspace{6pt} (21 topics) & 876 &
    Asset, Capital Asset Pricing Model, Dividend, Dividend payout ratio, Dividend yield, Earnings per share, Holding period return, Intrinsic value finance, Investments, Portfolios in finance, Preferred stock, Rate of return, Real gross domestic product, Residual income valuation, Risk premium, Security market line, Stock, Stocks, Stock exchange, Stock valuation, Valuation finance
    \\
    \addlinespace
    \textbf{2. Income \& Interest} (14 topics) & 618 &
    Bonds in finance, Compound interest, Coupon bond, Effective interest rate, Future value, Interest, Interest rate, International Fisher effect, Loan, Loans, Perpetuity, Present value, Yield to maturity, Zero-coupon bond
    \\
    \addlinespace
    \textbf{3. Financial Statements \& Analysis} \hspace{2pt} (10 topics) & 499 & Bad Debt, Balance sheets, Cash flow, Financial ratio, Financial statement analysis, Financial statements, Income statement, Interest expense, Inventory, Inventory turnover
    \\
    \addlinespace
    \textbf{4. Derivatives \& Risk Management} \hspace{2pt} (10 topics) & 433 & Bankruptcy, Credit, Derivative finance, Financial risk, Forward contract, Insurance, Interest rate risk, Options in finance, Risk management, Volatility Finance
    \\
    \addlinespace
    \textbf{5. Corporate Finance \& Capital Management} (11 topics) & 412 &
    Capital budgeting, Capital structure, Corporate finance, Cost of capital, Economic systems, Financial planning in business, Leverage in finance, Net present value, Payback period, Real options valuation, Weighted average cost of capital
    \\
    \addlinespace
    \textbf{6. Taxation \& Payroll} (7 topics) & 348 & 
    Corporate tax, Income tax, Payroll, Payroll tax, Progressive tax, Tax incidence, Taxes
    \\
    \addlinespace
    \textbf{7. Budgeting \& Personal Finance} \hspace{2pt} (9 topics) & 346 &
    Budget, Cash management, Checking account, Estate planning, Money management, Performance-based budgeting, Personal finance, Production budget, Savings account
    \\
    \bottomrule
  \end{tabularx}
  \caption{Seven finance categories and their 82 subtopics.}
  \label{tab:dataset_summary}
  \Description{Seven finance categories and their 82 subtopics}
\end{table}

\section{EXPERIMENT SETUP}
To evaluate the effectiveness of our framework, we conduct tests using out-of-the-box LLMs and a fine-tuned LLM on finance education datasets. 

\subsection{Question Sets}
We collected 3,532 questions from Study.com's finance curriculum~\footnote{The questions were obtained from \url{https://homework.study.com/learn/finance-questions-and-answers.html}. A subscription may be required to access the questions.}. The website contains hundreds of questions per topic, covering a spectrum of financial skills found in financial education such as university courses and professional preparatory materials like the CFA and CPA exams. We randomly collected 40-50 questions for each of the 82 finance topics. Each question includes the ground truth answer, a hint, and an expert explanation. We converted free-response questions into multiple-choice by using {GPT-o4-mini-high} to propose 3 alternative distractor options for the ground truth answer. As shown in Figure~\ref{fig:fcff_conversion}, we ensure the distractors are reasonably similar to the correct answer using Feng et al.'s prompting method~\cite{feng2024exploring}. Similarly, questions containing tables are converted into a linear text representation with bullet points denoting rows and columns that LLMs can easily read. 

\begin{figure}[h]
\centering
\begin{tcolorbox}[
  colback=gray!5,
  colframe=black,
  sharp corners,
  boxrule=0.6pt,
  arc=2pt,
  left=4pt,right=4pt,top=4pt,bottom=4pt
]
\small
\textbf{Topic:} Cash-Flow

\medskip\noindent
\textbf{Original free-response.}\par
\emph{Marlow Manufacturing reports the following figures for the most recent year (in thousands of USD):}

\begin{center}
\begin{tabular}{@{}l r@{}}
\toprule
\textbf{Item} & \textbf{Amount} \\ \midrule
Net income                              & \$140 \\ 
Depreciation \& amortization            & \$35  \\ 
Capital expenditures (gross PPE)        & \$60  \\ 
Change in net working capital (\(\Delta\)NWC) & \$18 \text{ (increase)} \\ 
After-tax interest expense              & \$12  \\ \bottomrule
\end{tabular}
\end{center}

Calculate the firm’s free cash flow to the firm (FCFF).

\emph{\textbf{Correct answer:}} \(\displaystyle \$109\,\text{ thousand}\).

\bigskip
\textbf{Converted to multiple choice}:\par

\emph{Marlow Manufacturing reports the following figures for the most recent year (in thousands of USD):}
\begin{itemize}
  \item Net income: \$140
  \item Depreciation \& amortization: \$35
  \item Capital expenditures (gross PPE): \$60
  \item Change in net working capital (\(\Delta\) NWC): \$18 (increase)
  \item After‑tax interest expense: \$12
\end{itemize}

What is the firm’s free cash flow to the firm (FCFF)?

\begin{enumerate}[label=(\Alph*)]
  \item \$\,91\,000
  \item \textbf{\$\,109\,000} \hfill \(\checkmark\)
  \item \$\,123\,000
  \item \$\,139\,000
\end{enumerate}

\end{tcolorbox}
\captionsetup{skip=0.5pt}
\caption{A tabular cash-flow free-response question: distractor answers are generated and the table is simplified into bullet points.}
\label{fig:fcff_conversion}
\Description{A tabular cash-flow free-response question: distractor answers are generated and the table is simplified into bullet points.}
\end{figure}

As shown in Table~\ref{tab:dataset_summary}, we group the 82 question sets into 7 finance categories: \emph{Investments \& Valuation}, \emph{Income \& Interest}, \emph{Financial Statements \& Analysis}, \emph{Derivatives \& Risk Management}, \emph{Corporate Finance \& Capital Management}, \emph{Taxation \& Payroll}, and \emph{Budgeting \& Personal Finance}.

\subsection{Implementation}

\subsubsection*{\normalfont\bfseries Crafting Role Prompts}
Liu et al. showed using LLM-simulated role prompts during ideation significantly enhances user-perceived quality of outcomes such as the relevance of critiques~\cite{liu2024personaflow}. LLM-generated role prompts obtained more stable and relevant results than handcrafted ones, reducing performance variance~\cite{kim2024persona}. The Expert Reviewer loads a unique system role prompt for each of the 82 subtopics. Figure~\ref{tab:role_prompts} shows a sample of the topic and role prompts. We use {GPT-o4-mini-high} to generate each system prompt by defining an expert persona (``You are a(n) \_ expert") and 2) including one to two domain-specific responsibilities (Figure~\ref{tab:role_prompts}). After, each role prompt is reviewed via a human pass to ensure correctness and relevance to the role. 

For OpenAI models to support explicit system prompt definition via the Chat Completions API, we provide each topic role as the {system} message. For other models without a dedicated system prompt, we prepend the topic to the front of the question query. This enables all models tested to receive the same role context. 

\subsubsection*{\normalfont\bfseries Retrieval-Augmented Generation (RAG)}
To build our RAG knowledge base, we indexed 6 well-established finance textbooks~\cite{palepu2013business,hull2012options,brealey2019principles,horngren2011financial,tuckman2002fixed,bodie2013investments}:
\begin{itemize}
  \item \emph{Business Analysis and Valuation}, Palepu, Healy, and Peek. 
  \item \emph{Options, Futures, and Other Derivatives}, Hull.
  \item \emph{Principles of Corporate Finance}, Brealey, Myers, and Allen.
  \item \emph{Financial \& Managerial Accounting}, Prentice Hall.
  \item \emph{Fixed Income Securities}, Tuckman, Wiley Finance.
  \item \emph{Investments}, Bodie, Kane, and Marcus.
\end{itemize}
These textbooks were split into passage-level chunks and embedded using {all-MiniLM-L6-v2} model. At query time, we retrieve the top-3 most semantically similar passages.   

\subsection{Baseline Methods} 
We test our framework on 3 proprietary and state-of-the-art models - {GPT-4o-mini}, {Gemini-2.0-Flash}, and {Llama-3.1-70B-Instruct}. {GPT-4o-mini} is fast in reasoning with low inference costs. {Gemini-2.0-Flash} is a closed-source alternative with strong capacities for handling complex data. {Llama-3.1-70B-Instruct} is a large, open-weight model fine-tuned for instruction following. Finally, we evaluate {FinGPT-mt\_Llama3-8B\_LoRA}, the latest FinGPT model. We compare against out-of-the-box and fine-tuned finance models (serving as performance upper bounds) where we can quantify the gains introduced by our framework. The temperature is set to 0.1 across all models to ensure outputs are consistent for the same question. The output is set to a maximum of 1000 tokens. 

\footnotetext[1]{\url{https://huggingface.co/FinGPT/fingpt-mt_llama3-8b_lora}} 

\begin{figure*}[t] 
  \centering 
  \scriptsize 

\begin{subfigure}[t]{0.44\textwidth}
  \centering
  \scriptsize
  \begin{tikzpicture}
    \begin{axis}[
      ybar,
      axis lines=box,
      bar width=4pt,
      width=\columnwidth,
      height=5.7cm,
      enlarge x limits=0.1,
      ymin=40,  
      ymax=90,
      ytick distance=10,
      ylabel={Accuracy (\%)},
      symbolic x coords={
        {Invest. \& Val},
        {Income \& Interest},
        {Fin. Stmts. \& Analysis},
        {Deriv. \& Risk Mgmt},
        {Corp. Fin. \& Cap. Mgmt},
        {Tax \& Payroll},
        {Budgeting \& Pers. Finance}
      },
      xtick=data,
      x tick label style={
        rotate=25,
        anchor=east,
        font=\scriptsize,
        text width=2.2cm,
        align=right, 
        yshift=-3pt,
        /pgf/number format/fixed,
        /pgf/number format/precision=1,
        /pgf/number format/zerofill
      },
      xtick pos=bottom,
      ytick pos=left,
      nodes near coords,
      every node near coord/.append style={
        font=\fontsize{4.5}{4.8}\selectfont,
        color=black!85,
        rotate=90,
        anchor=west,
        /pgf/number format/fixed,
        /pgf/number format/precision=1,
        /pgf/number format/zerofill
      },
      legend style={ 
        legend columns=4, 
        at={(1.85,1.2)},
        anchor=north east,
        font=\small,
        draw=black!60,
        fill=white,
        legend cell align=left,
        /tikz/every even column/.append style={column sep=4pt}
      },
      legend image code/.code={
        \draw[#1, rounded corners=1pt]
        (0.0cm,-0.075cm) rectangle (0.3cm,0.075cm);
      },
      legend image post style={
        draw=black!60,
        fill opacity=1.0,
        inner sep=0pt,
        outer sep=1pt
      }
    ]   
    \addplot+[bar shift=-6pt, fill=finGray, draw=black!40] coordinates {
        (Invest. \& Val,64.1)
        (Income \& Interest,63.3)
        (Fin. Stmts. \& Analysis,65.8)
        (Deriv. \& Risk Mgmt,61.2)
        (Corp. Fin. \& Cap. Mgmt,65.2)
        (Tax \& Payroll,64.2)
        (Budgeting \& Pers. Finance,59.6)
    };
    \addlegendentry{M-0 (Baseline)}
    
    \addplot+[bar shift=-2pt, fill=finEv, draw=black!40] coordinates {
        (Invest. \& Val,65.5)
        (Income \& Interest,65.2)
        (Fin. Stmts. \& Analysis,67.7)
        (Deriv. \& Risk Mgmt,63.0)
        (Corp. Fin. \& Cap. Mgmt,66.3)
        (Tax \& Payroll,65.6)
        (Budgeting \& Pers. Finance,60.8)
    };
    \addlegendentry{M-1 (+Evidence)}
    
    \addplot+[bar shift=2pt, fill=finCr, draw=black!40] coordinates {
        (Invest. \& Val,67.0)
        (Income \& Interest,66.6)
        (Fin. Stmts. \& Analysis,70.2)
        (Deriv. \& Risk Mgmt,65.6)
        (Corp. Fin. \& Cap. Mgmt,69.0)
        (Tax \& Payroll,68.1)
        (Budgeting \& Pers. Finance,63.3)
    };
    \addlegendentry{M-2 (+Critique)}
    
    \addplot+[bar shift=6pt, fill=finFull, draw=black!40] coordinates {
        (Invest. \& Val,70.4)
        (Income \& Interest,70.1)
        (Fin. Stmts. \& Analysis,72.8)
        (Deriv. \& Risk Mgmt,68.2)
        (Corp. Fin. \& Cap. Mgmt,71.7)
        (Tax \& Payroll,70.3)
        (Budgeting \& Pers. Finance,66.0)
    };
      \addlegendentry{M-3 (Full)}
    \end{axis}  
  \end{tikzpicture}
  \vspace{-9pt}
  \caption{GPT-4o-mini}
  \label{fig:grouped_bars_nogap_1}
  \Description{GPT-4o-mini results}
\end{subfigure}
  \hspace{0.04\textwidth}
  \begin{subfigure}[t]{0.44\textwidth}
          \centering
      \scriptsize
      \begin{tikzpicture}
        \begin{axis}[
          ybar,
          axis lines=box, 
          bar width=4pt,
          width=\columnwidth,
          height=5.7cm,
          enlarge x limits=0.1,
          ymin=40,  
          ymax=90,
          ytick distance=10,
          ylabel={Accuracy (\%)},
          symbolic x coords={
            {Invest. \& Val},
            {Income \& Interest},
            {Fin. Stmts. \& Analysis},
            {Deriv. \& Risk Mgmt},
            {Corp. Fin. \& Cap. Mgmt},
            {Tax \& Payroll},
            {Budgeting \& Pers. Finance}
          },
          xtick=data,
                x tick label style={
        rotate=25,
        anchor=east,
        font=\scriptsize,
        text width=2.2cm,
        align=right, 
        yshift=-3pt,
        /pgf/number format/fixed,
        /pgf/number format/precision=1,
        /pgf/number format/zerofill
      },
      xtick pos=bottom,
      ytick pos=left,
      nodes near coords,
      every node near coord/.append style={
        font=\fontsize{4.5}{4.8}\selectfont,
        color=black!85,
        rotate=90,
        anchor=west,
        /pgf/number format/fixed,
        /pgf/number format/precision=1,
        /pgf/number format/zerofill
      },
        ]
        \addplot+[bar shift=-6pt, fill=finGray, draw=black!40] coordinates {
            (Invest. \& Val,73.1)            
            (Income \& Interest,71.7)
            (Fin. Stmts. \& Analysis,75.5)
            (Deriv. \& Risk Mgmt,66.6)
            (Corp. Fin. \& Cap. Mgmt,74.8)
            (Tax \& Payroll,70.4)
            (Budgeting \& Pers. Finance,64.0)
        };
        
        \addplot+[bar shift=-2pt, fill=finEv, draw=black!40] coordinates {
            (Invest. \& Val,75.2)            
            (Income \& Interest,73.3)        
            (Fin. Stmts. \& Analysis,77.7)   
            (Deriv. \& Risk Mgmt,68.7)       
            (Corp. Fin. \& Cap. Mgmt,76.1)   
            (Tax \& Payroll,72.3)            
            (Budgeting \& Pers. Finance,66.1)
        };
        
        \addplot+[bar shift=2pt, fill=finCr, draw=black!40] coordinates {
            (Invest. \& Val,79.3)            
            (Income \& Interest,77.3)        
            (Fin. Stmts. \& Analysis,81.9)   
            (Deriv. \& Risk Mgmt,72.7)       
            (Corp. Fin. \& Cap. Mgmt,80.5)   
            (Tax \& Payroll,76.5)            
            (Budgeting \& Pers. Finance,70.5)
        };
        
        \addplot+[bar shift=6pt, fill=finFull, draw=black!40] coordinates {
            (Invest. \& Val,81.3)            
            (Income \& Interest,79.4)        
            (Fin. Stmts. \& Analysis,84.1)   
            (Deriv. \& Risk Mgmt,74.7)       
            (Corp. Fin. \& Cap. Mgmt,82.7)   
            (Tax \& Payroll,78.8)            
            (Budgeting \& Pers. Finance,73.0)
        };
        \end{axis}
      \end{tikzpicture}
      \vspace{-8pt}
      \caption{Gemini-2.0-Flash}
      \label{fig:grouped_bars_nogap_2}
      \Description{Gemini-2.0-Flash results}
  \end{subfigure}
  \begin{subfigure}[t]{0.44\textwidth}
          \centering
      \scriptsize
      \begin{tikzpicture}
        \begin{axis}[
          ybar,
          axis lines=box, 
          bar width=4pt,
          width=\columnwidth,
          height=5.7cm,
          enlarge x limits=0.1,
          ymin=40,  
          ymax=90,
          ytick distance=10,
          ylabel={Accuracy (\%)},
          symbolic x coords={
            {Invest. \& Val},
            {Income \& Interest},
            {Fin. Stmts. \& Analysis},
            {Deriv. \& Risk Mgmt},
            {Corp. Fin. \& Cap. Mgmt},
            {Tax \& Payroll},
            {Budgeting \& Pers. Finance}
          },
         x tick label style={
            rotate=25,
            anchor=east,
            font=\scriptsize,
            text width=2.2cm,
            align=right, 
            yshift=-3pt,
            /pgf/number format/fixed,
            /pgf/number format/precision=1,
            /pgf/number format/zerofill
          },
          xtick pos=bottom,
          ytick pos=left,
          nodes near coords,
          every node near coord/.append style={
            font=\fontsize{4.5}{4.8}\selectfont,
            color=black!85,
            rotate=90,
            anchor=west,
            /pgf/number format/fixed,
            /pgf/number format/precision=1,
            /pgf/number format/zerofill
          },
        ]
        \addplot+[bar shift=-6pt, fill=finGray, draw=black!40] coordinates {
            (Invest. \& Val,62.3) 
            (Income \& Interest,61.4)
            (Fin. Stmts. \& Analysis,64.1)
            (Deriv. \& Risk Mgmt,59.7)
            (Corp. Fin. \& Cap. Mgmt,62.9)
            (Tax \& Payroll,61.7)
            (Budgeting \& Pers. Finance,57.4)
        };
        
        \addplot+[bar shift=-2pt, fill=finEv, draw=black!40] coordinates {
            (Invest. \& Val,63.7)
            (Income \& Interest,63.3)
            (Fin. Stmts. \& Analysis,65.8)
            (Deriv. \& Risk Mgmt,61.1)
            (Corp. Fin. \& Cap. Mgmt,64.7)
            (Tax \& Payroll,63.5)
            (Budgeting \& Pers. Finance,58.8)
        };
        
        \addplot+[bar shift=2pt, fill=finCr, draw=black!40] coordinates {
            (Invest. \& Val,65.8)
            (Income \& Interest,66.4)
            (Fin. Stmts. \& Analysis,69.2)
            (Deriv. \& Risk Mgmt,62.5)
            (Corp. Fin. \& Cap. Mgmt,67.8)
            (Tax \& Payroll,66.7)
            (Budgeting \& Pers. Finance,61.4)
        };
        
        \addplot+[bar shift=6pt, fill=finFull, draw=black!40] coordinates {
            (Invest. \& Val,69.0)
            (Income \& Interest,68.9)
            (Fin. Stmts. \& Analysis,71.2)
            (Deriv. \& Risk Mgmt,66.3)
            (Corp. Fin. \& Cap. Mgmt,70.0)
            (Tax \& Payroll,68.9)
            (Budgeting \& Pers. Finance,64.7)
        };
        \end{axis}
      \end{tikzpicture}
      \vspace{-3pt}
    \caption{Llama-3.1-70B-Instruct}
    \label{fig:grouped_bars_nogap_3}
    \Description{Llama-3.1-70B-Instruct}
  \end{subfigure}%
  \hspace{0.04\textwidth}%
  \begin{subfigure}[t]{0.44\textwidth}
          \centering
      \scriptsize
      \begin{tikzpicture}
        \begin{axis}[
          ybar,
          axis lines=box, 
          bar width=4pt,
          width=\columnwidth,
          height=5.7cm,
          enlarge x limits=0.1,
          ymin=40,  
          ymax=90,
          ytick distance=10,
          ylabel={Accuracy (\%)},
          symbolic x coords={
            {Invest. \& Val},
            {Income \& Interest},
            {Fin. Stmts. \& Analysis},
            {Deriv. \& Risk Mgmt},
            {Corp. Fin. \& Cap. Mgmt},
            {Tax \& Payroll},
            {Budgeting \& Pers. Finance}
          },
        x tick label style={
            rotate=25,
            anchor=east,
            font=\scriptsize,
            text width=2.2cm,
            align=right, 
            yshift=-3pt,
            /pgf/number format/fixed,
            /pgf/number format/precision=1,
            /pgf/number format/zerofill
          },
          xtick pos=bottom,
          ytick pos=left,
          nodes near coords,
          every node near coord/.append style={
            font=\fontsize{4.5}{4.8}\selectfont,
            color=black!85,
            rotate=90,
            anchor=west,
            /pgf/number format/fixed,
            /pgf/number format/precision=1,
            /pgf/number format/zerofill
          },
        ]

    \addplot+[bar shift=-6pt, fill=finGray, draw=black!40] coordinates {
            (Invest. \& Val,65.7)
            (Income \& Interest,65.6)
            (Fin. Stmts. \& Analysis,68.5)
            (Deriv. \& Risk Mgmt,62.6)
            (Corp. Fin. \& Cap. Mgmt,67.0)
            (Tax \& Payroll,65.8)
            (Budgeting \& Pers. Finance,61.3)
        };
    
    \addplot+[bar shift=-2pt, fill=finEv, draw=black!40] coordinates {
        (Invest. \& Val,67.7)         
        (Income \& Interest,66.8)     
        (Fin. Stmts. \& Analysis,69.5)
        (Deriv. \& Risk Mgmt,64.8)    
        (Corp. Fin. \& Cap. Mgmt,68.7)
        (Tax \& Payroll,67.6)         
        (Budgeting \& Pers. Finance,62.7)
    };
    
    \addplot+[bar shift=2pt, fill=finCr, draw=black!40] coordinates {
        (Invest. \& Val,70.9)            
        (Income \& Interest,70.0)        
        (Fin. Stmts. \& Analysis,72.8)   
        (Deriv. \& Risk Mgmt,68.2)       
        (Corp. Fin. \& Cap. Mgmt,71.9)   
        (Tax \& Payroll,71.1)            
        (Budgeting \& Pers. Finance,66.3)
    };
    
    \addplot+[bar shift=6pt, fill=finFull, draw=black!40] coordinates {
        (Invest. \& Val,73.5)            
        (Income \& Interest,73.3)       
        (Fin. Stmts. \& Analysis,74.2)   
        (Deriv. \& Risk Mgmt,70.9)       
        (Corp. Fin. \& Cap. Mgmt,74.7)   
        (Tax \& Payroll,74.1)            
        (Budgeting \& Pers. Finance,69.2)
    };
        \end{axis}
      \end{tikzpicture}
      \vspace{-3pt}
    \caption{FinGPT-mt\_Llama3-8B\_LoRA}
    \label{fig:grouped_bars_nogap_4}
    \Description{FinGPT-mt\_Llama3-8B\_LoRA results}
  \end{subfigure}
  \caption{Performance breakdown by financial category across 4 prompting modes (M-0 to M-3) for GPT-4o-mini, Gemini-2.0-Flash, Llama-3.1-70B-Instruct, and FinGPT-mt\_Llama3-8B\_LoRA}.
  \label{fig:all_models_results}
  \Description{Performance breakdown by financial category across 4 prompting modes (M-0 to M-3) for GPT-4o-mini, Gemini-2.0-Flash, Llama-3.1-70B-Instruct, and FinGPT-mt\_Llama3-8B\_LoRA}
\end{figure*}
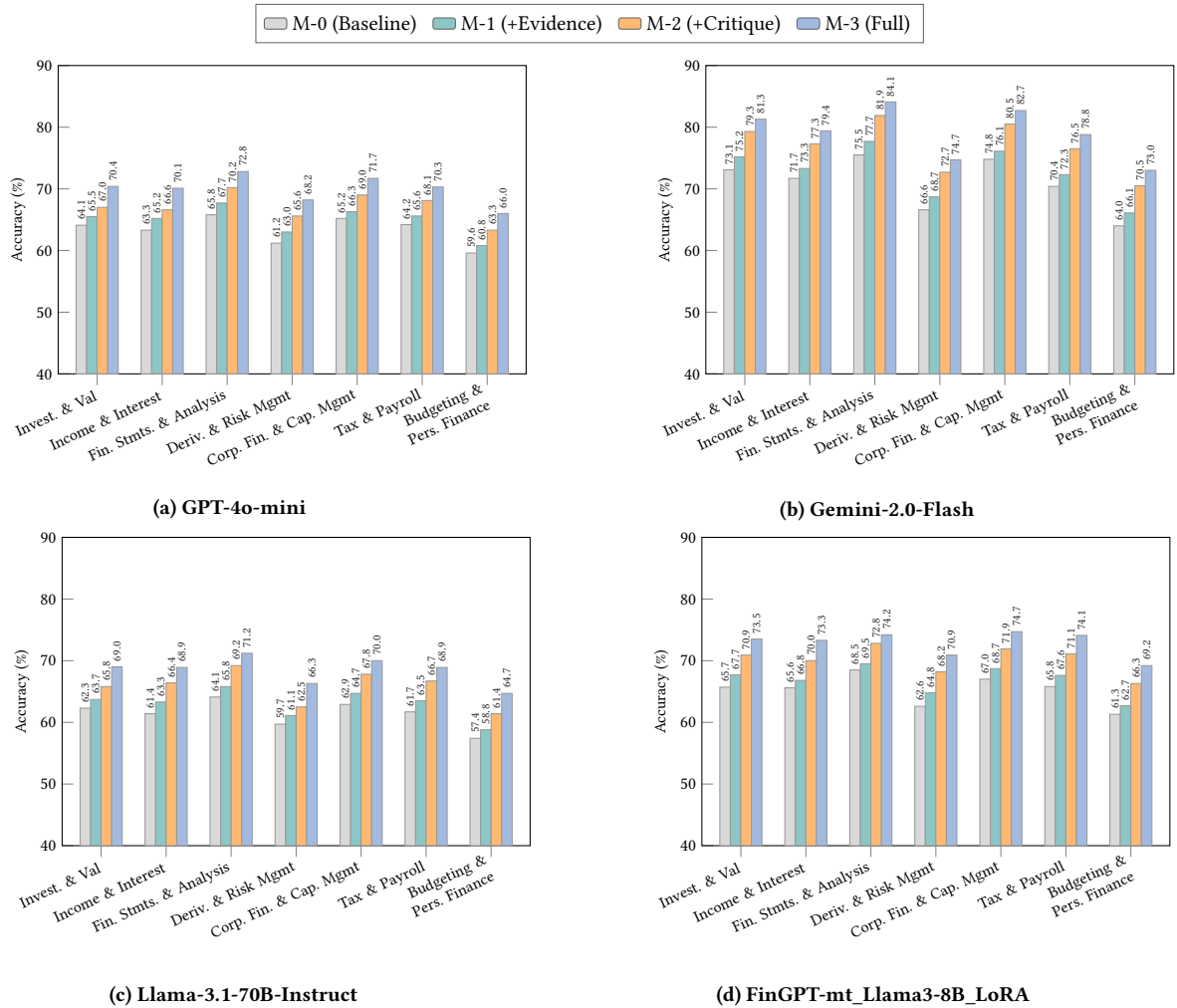

\begin{table*}[t]
  \centering
  \setlength{\tabcolsep}{8pt}
  \begin{tabular}{lccccc}
    \toprule
    \textbf{Model} & \textbf{M-0 (Baseline)} & \textbf{M-1 (+Evidence)} & \textbf{M-2 (+Critique)} & \textbf{M-3 (Full)} & \textbf{Gain} \\
    \midrule
    GPT-4o-mini             & 63.34 & 64.87 & 67.11 & 69.93 & \textbf{+6.59} \\
    Gemini-2.0-Flash        & 70.87 & 72.77 & 76.96 & 79.14 & \textbf{+8.27} \\
    Llama-3.1-70B-Instruct  & 61.36 & 62.99 & 65.69 & 68.43 & \textbf{+7.07} \\
    \midrule
    FinGPT-mt\_Llama3-8B\_LoRA & 65.21 & 66.83 & 70.17 & 72.84 & \textbf{+7.63} \\
    \bottomrule
  \end{tabular}
  \captionsetup{width=0.95\linewidth}
  \caption{Average accuracy of each model under the M-0 through M-3 configurations. The rightmost column reports overall gain (M-3 - M-0).}
  \label{tab:model_performance}
\end{table*}

\section{Results}

\subsection{Performance Evaluation}
\noindent{\textbf{Evaluation Metric.}}
Model performance is calculated as the proportion of questions that the model's final choice matches the ground-truth multiple choice option (A-D). 
Figure~\ref{fig:all_models_results} shows the evaluation results across the question sets and Table~\ref{tab:model_performance} summarizes the overall average accuracy across the 4 configurations (M-0 through M-3). The framework consistently boosts accuracy at each stage, with particularly stronger gains from the Expert Reviewer agent. 

\noindent\textbf{Closed-book (M-0).} In the closed-book setting, all models rely on zero-shot Chain-of-Thought reasoning. {Gemini-2.0-Flash} leads at 70.87\%, followed by FinGPT at 65.21\%, {GPT-4o-mini} at 63.34\%, and {Llama-3.1-70B-Instruct} at 61.36\%. The gap between proprietary models like {Gemini-2.0-Flash} and open-source models like {Llama-3.1-70B-Instruct} underscores the advantages of larger pre-training. {FinGPT-mt\_Llama3-8B-LoRA} also highlights targeted fine-tuning with low-rank adaptation on multi-task financial corpus can inject domain expertise into smaller models and outperform larger models like {Llama-3.1-70B-Instruct}. 

\noindent\textbf{Impact of Evidence Retrieval (M-1).} Adding relevant textbook evidence yields a slight improvement for each model, with {GPT-4o-mini} climbing +1.53\%, {Gemini-2.0-Flash} climbing +1.90\%, Llama-3.1-70B-Instruct climbing +1.63\%, and FinGPT climbing +1.62\%. This suggests that evidence retrieval helps with question-answering but is not the dominant driver in reducing hallucinations and filling knowledge gaps.
However, outputs from the Evidence Retriever agent have the most profound performance impact when they are concept-heavy or contain under-specified questions.

\noindent\textbf{Impact of Critique (M-2).} 
Using the reviewer's critique drives substantial improvements in performance. It has shown to influence the model's second pass output, attributing to +2.24\% for {GPT-4o-mini}, +4.19\% for {Gemini-2.0-Flash}, +2.70\% for {Llama-3.1-70B-Instruct}, and +3.34\% for FinGPT. Our observations into the critique output find that the agent detects arithmetic errors and wrong steps, and it outputs constructive feedback for the base agent to consider. This shows that critique agent corrects mistakes more than evidence alone.  

\noindent\textbf{Combined Performance (M-3).} 
Combining both retrieval and critique into the QA framework yields the highest performance across all models. This configuration gains +6.59\% for {GPT-4o-mini}, +8.27\% for {Gemini-2.0-Flash}, +7.07\% for {Llama-3.1-70B-Instruct}, and +7.63\% for FinGPT. The larger jump in M-3 performance can be explained by evidence removing factual errors while critique corrects procedural mistakes; reducing both error sources produces compounded improvement. Together, the combined work of the evidence retriever and reviewer agents enforces factual accuracy and arithmetic consistency.  

Upon inspecting initially incorrect responses, the Base Generator agent is able to successfully incorporate feedback from the evidence and reviwer agents. In the second fresh-context, it mostly (i) recomputes intermediate values, (ii) uses the correct formulas/terminologies, and (iii) produces more confident and structured step-by-step derivations. 

Within the seven finance categories, Budgeting \& Personal Finance was the most difficult category for the models as it mixes everyday language with financial terminologies, and it requires translating broad contexts into multi-step arithmetic. This category also shows some of the largest gains to M-3 (e.g., Gemini: +9.0; FinGPT: +7.9), indicating retrieval clarifies definitions while expert critique fixes arithmetic. 

\subsection{Fine-Tuned Baseline Comparison} 
We find that our pipeline can elevate proprietary models to rival and exceed the performance of FinGPT model without fine-tuning or adjusting model weights. {GPT-4o-mini's} average M-3 accuracy of 69.9 surpasses FinGPT's zero-shot baseline of 65.2\%, while Llama-3.1-70B-Instruct's average of 68.4\% outperforms that same FinGPT baseline by 3.2\%. FinGPT itself also gains substantial improvements from evidence retrieval and critique, climbing from 65.2\% to 72.9\%. We demonstrate our pipeline can compensate for smaller fine-tuning budgets and open-source constraints.

\subsection{Cost Analysis}
Pre-training and fine-tuning can be expensive as they require heavy data preparation and extensive resources. Our agent framework achieves higher performance on financial QA by reusing cost-efficient models like {GPT-4o-mini} and {Gemini-2.0-Flash} during the agent calls. Each mode adds one extra LLM call where accuracy rises with linear cost rather than from inferring larger models: M-0 (1 call), M-1 (2), M-2 (3), M-3 (4). For RAG, the textbooks are cached in vector embeddings, making evidence retrieval calls also relatively inexpensive. 

We ran the open-weight models ({Llama-3.1-70B-Instruct} and {FinGPT-mt\_Llama3-8B\_LoRA}) on an A100 cloud GPU. The 2 proprietary models were called through APIs. The experiments for {GPT-4o-mini} cost $\approx$ \$65 end-to-end. Exact billing usage for {Gemini-2.0-Flash} was not available to us, but we estimate it to be slightly lower than the GPT model due to lower token costs. 


\section{CONCLUSION}
In financial education, mastery implies the ability to reason through financial QA. Our work presents a role-aware multi-agent framework for enhancing LLM performance on financial QA. Benchmarks on 3,532 finance education-level questions show our framework yields accuracy gains over zero-shot Chain-of-Thought, with the largest improvements coming from the critique supplied by the reviewer agent. The results show that evidence and reviewer agents work hand-in-hand to reinforce models with textbook-backed definitions and correct mistakes during multi-step reasoning. RAG can help steer LLMs to factual accuracy, and a role-aware LLM can serve as a critic agent to correct a model's output with significant improvements. Performance insights also reveal models struggle the most with derivatives and risk management questions, reinforcing the need for attention in topics like financial risk, credit, and volatility. Overall, we show that proprietary models can achieve competitive performance similar to the finance fine-tuned FinGPT without extra costs and computational resources. We demonstrate that structured agent collaboration can be an effective alternative to expensive model training and tuning for finance education. 

In the future, we will investigate how the structure of critique impacts performance to better craft system prompts. In addition, we will explore external tools such as Program-of-Thoughts, calculators, and table parsers to ground reasoning steps and reduce errors. We also plan to benchmark our framework on weaker topics found in our analysis, as well as expand our testing data to other finance education sources. 

\bibliographystyle{ACM-Reference-Format} 
\bibliography{main}

\onecolumn

\end{document}